# Democratising Artificial Intelligence for Pandemic Preparedness and Global Governance in Latin American and Caribbean Countries

*André de Carvalho[1,15*], Robson Bonidia[1,15*], Jude Dzevela Kong[15,16*], Mariana Dauhajre[12,15*], Claudio Struchiner[3,15], Guilherme Goedert[3,15], Peter F. Stadler[4,15], Maria Emilia Walter[5,15], Danilo Sanches[6,15], Troy Day[7,15], Marcia Castro[8,15], John Edmunds[9,15], Manuel Colomé-Hidalgo[10,15], Demian Arturo Herrera Morban[11,15], Edian F. Franco[13,14,15],* Cesar Ugarte-Gil[15,17], Patricia Espinoza-Lopez[15,17], Gabriel Carrasco-Escobar[15,17], *Ulisses Rocha[2,15*]*

[1]*Institute of Mathematics and Computer Sciences, University of São Paulo, São Carlos 13566-590, Brazil;* [2]*Department of Environmental Microbiology, Helmholtz Centre for Environmental Research-UFZ GmbH, 04318 Leipzig, Germany;* [3]*Escola de Matemática Aplicada, Fundação Getúlio Vargas, Rio de Janeiro, Brazil;* [4]*Department of Computer Science and Interdisciplinary Center of Bioinformatics, University of Leipzig, 04107 Leipzig, Germany;* [5]*Department of Computer Science, University of Brasília, Brasília, Brazil;* [6]*Computer Science Department, Federal University of Technology - Paraná, Brazil;* [7]*Professor and Head, Department of Mathematics and Statistics, Queen's University, Canada;* [8]*Harvard T.H. Chan School of Public Health, 665 Huntington Avenue, Bldg I, Room 1102A. Boston, MA;* [9]*London School of Hygiene and Tropical Medicine, United Kingdom;* [10]*Professor and Leader, Instituto de Investigación en Salud, Universidad Autónoma de Santo Domingo, Dominican Republic;* [11]*Research Coordinator, Centro de Investigación en Salud Dr. Hugo Mendoza, Hospital Pediátrico Dr. Hugo Mendoza, Santo Domingo, Dominican Republic;* [12]*Research Fellow, Centro de Investigación en Salud Dr. Hugo Mendoza, Hospital Pediátrico Dr. Hugo Mendoza, Santo Domingo, Dominican Republic;* [13]*Centro de Investigación UTESA, Universidad Tecnologica de Santiago (UTESA);* [14]*School of Basic and Environmental Sciences, Instituto Tecnologico de Santo Domingo (INTEC);* [15]*Global South Artificial Intelligence for Pandemic and Epidemic Preparedness and Response Network (AI4PEP);* [16]*Department of Mathematics and Statistics York University, N520 Ross, 4700 Keele Street, Toronto, ON M3J 1P3;* [17]*Instituto de Medicina Tropical Alexander von Humboldt, Universidad Peruana Cayetano Heredia, Av. Honorio Delgado 430, San Martín de Porres 15102, Peru*

*These authors contributed equally.
**Corresponding authors:** andre@icmc.usp.br; rpbonidia@gmail.com; ulisses.rocha@ufz.de

**Abstract:** Infectious diseases, transmitted directly or indirectly, are among the leading causes of epidemics and pandemics. Consequently, several open challenges exist in predicting epidemic outbreaks, detecting variants, tracing contacts, discovering new drugs, and fighting misinformation. Artificial Intelligence (AI) can provide tools to deal with these scenarios, demonstrating promising results in the fight against the COVID-19 pandemic. AI is becoming increasingly integrated into various aspects of society. However, ensuring that AI benefits are distributed equitably and that they are used responsibly is crucial. Multiple countries are creating regulations to address these concerns, but the borderless nature of AI requires global cooperation to define regulatory and guideline consensus. Considering this, The Global South AI for Pandemic & Epidemic Preparedness & Response Network (AI4PEP) has developed an initiative comprising 16 projects across 16 countries in the Global South, seeking to strengthen equitable and responsive public health systems that leverage Southern-led responsible AI solutions to improve prevention, preparedness, and response to emerging and re-emerging infectious disease outbreaks. This opinion introduces our branches in Latin American and Caribbean (LAC) countries and discusses AI governance in LAC in the light of biotechnology. Our network in LAC has high potential to help fight infectious diseases, particularly in low- and middle-income countries, generating opportunities for the widespread use of AI techniques to improve the health and well-being of their communities.

## 1. Introduction and Problem Statement

Lessons learned from the COVID-19 pandemic outbreak highlight the need to improve our preparedness for similar events [1]. This pandemic has shown us that no single person, community, or country is isolated and that suffering and lack of support affecting single individuals, independent of where they are, affect all of us. Moreover, we must recognise talent, wherever and whoever they are, to join and collaborate in the search for solutions to the world's challenges.

In this effort, Artificial Intelligence (AI) emerges as a valuable tool to reduce pandemic impacts [2]. In an era where AI is present in various processes that impact society, it is essential to ensure that its contributions are distributed equitably [3]. Moreover, the democratisation of AI must empower each individual, community, or society to contribute proportionately to their aptitude, availability, dedication, and speed [4].

With AI's advances, concerns, and contributions, the world has received positive and negative signals regarding the future of a world where AI is firmly present [5]. Advances in all knowledge domains show how AI can not only accelerate scientific discoveries and the design of innovative solutions, many of them improving people's health, but also be one of the most valuable tools for improving the quality of life on earth [6]. On the other hand, recent articles, talks, and interviews address the dangers of developing and using AI [5, 7, 8].

Law-making bodies in many countries and regions are passing new regulations regarding the development and use of AI to reduce or avoid risks associated with this technology [9, 10]. These regulations may prevent biased AI systems, ensure fairness, and improve privacy protection. Nevertheless, AI has no borders. If one country is rigid and another is permissive on the same issue, harmful efforts will be directed to the more permissive country. Recent articles have defended the idea that AI regulation should not be the responsibility or work of a single country or region. Therefore, a global board should indicate the best actions to create legislation regulating AI, similar to the Intergovernmental Panel on Climate Change (IPCC), an intergovernmental body of the United Nations [11, 12].

The democratisation of AI aims at its ethical and responsible expansion, which involves its development, use, and control [4, 13]. These aims require equal opportunities across the world. Furthermore, we should lay plans to share knowledge and training involving AI to benefit all regions of the globe [13].

Considering these issues, York University (Toronto, Canada) was selected in a Canadian national bid for a grant from the International Development Research Centre (IDRC) (AI and Global Health Investment[1]) to support countries in the global south to fight infectious diseases using AI, The Global South AI for Pandemic & Epidemic Preparedness & Response Network (AI4PEP[2]). This network comprises 16 projects from 16 countries in the global south. This paper introduces our branches in Latin American and Caribbean (LAC) countries and discusses AI governance in LAC.

---

[1]https://idrc-crdi.ca/en/initiative/artificial-intelligence-global-health
[2]https://ai4pep.org/

The Brazilian branch of the network, AutoAI-Pandemics[3], will investigate and design automated Machine Learning (ML) tools to democratise healthcare professionals' use of AI. These tools will support (1) automated epidemiological analysis for designing interventions, (2) automated bioinformatics analysis, and (3) fighting misinformation/disinformation. AutoAI-Pandemics hopes to democratise access to data science and ML techniques by allowing non-experts (e.g., biologists, physicians, and epidemiologists) to use AI in their research and development.

The Dominican branch, AI4EWARS[4], plans to use AI to model the spread of Aedes-borne diseases in the country. It comprises multidisciplinary and experienced health researchers, including entomologists, mathematicians, and medical doctors. They plan to design a predictive model of Aedes-borne viruses in the Dominican Republic that provides early warnings of outbreaks and streamlines Public Health responses.

The AI4PEP hub in Peru will enhance an existing database of forced cough sounds to further train pilot AI algorithms for classifying various respiratory infectious diseases in patients with respiratory symptoms. The project aims to evaluate the utility of longitudinal cough monitoring to index patients and their household contacts and to integrate AI-based tools to improve respiratory infection surveillance, access, and equity within the Peruvian healthcare system.

Our network in LAC has the potential to significantly reduce the experience needed to use ML pipelines, helping researchers in the fight against infectious diseases, mainly in low—and middle-income countries. This initiative will allow biologists, physicians, epidemiologists, and other stakeholders to use these techniques widely to improve the health and well-being of their communities.

## 2. Challenges and Inequalities in LAC

AI solutions have been proposed in several domains (e.g., healthcare, finances, education, agriculture). In October 2022, the Food and Drug Administration (FDA) reported 521 AI and ML-enabled medical devices [14]. Furthermore, the COVID-19 pandemic demonstrated the potential of ML techniques to minimise the effects of a pandemic, such as the prediction of deaths, contact tracing, diagnosis, treatments, and others [15, 16].

---

[3]Official Page: autoaipandemics.icmc.usp.br/
[4]Official Page: https://ai4ewars.org/

Nevertheless, many studies have a black-box nature (i.e., AI decisions are not understandable on a human level) [17, 18], which may reduce AI's trust, accountability, and acceptance [19]. Another concern is that ML models can follow hidden social biases in the data, leading to unfair, harmful, or prejudiced decisions. Some examples of these problems were recently reported. In 2009, genome-wide association studies had more than 96% of participants of European descent [20], failing to encompass racial and geographic diversity. Cirillo and collaborators [21] reported sex and gender differences and biases in AI for biomedicine and healthcare. In dermatology, medical imaging, and diabetes management, studies have discussed the lack of racial diversity in ML algorithms, with possible risks of health disparities [22, 23, 39].

Consequently, initiatives have discussed concerns regarding ethical, fair, reliable, sustainable, transparent, and reproducible AI [24]. Thus, in search of responsible solutions, our network follows guidelines proposed in the literature, such as how to develop and use AI responsibly [25], AI for all [26], Guidelines for Trustworthy AI [27], Ethics of AI (UNESCO), and others. We also adopt and recommend the principles of Data-Centric AI [28], putting data at the heart of an AI system development process, and the guidelines to improve the Findability, Accessibility, Interoperability, and Reuse of data known as FAIR data principles [29].

## 3. Comparative AI Governance Models in LAC

According to a report by the OECD[5], in 2022 [30], AI capacities in LAC countries vary widely. Countries like Trinidad and Tobago, Venezuela, and Bolivia have yet to achieve significant public sector development. Uruguay and Colombia are regional leaders in AI strategy research and development. These countries are followed by Peru, Chile, Brazil, Costa Rica, and Argentina, which have drafted or enacted legislation about AI. In addition, Mexico was the first LAC nation to create a national strategy for AI, as shown in Table 1. Lastly, Paraguay, Panama, Jamaica, Ecuador,

---

[5] https://oecd-opsi.org/publications/ai-lac/

the Dominican Republic, and Barbados are beginning to integrate public sector strategies for AI.

**Table 1: AI laws and regulations in different Latin American Countries**

| Year | Country | AI Definition | Compare and Contrast |
|---|---|---|---|
| 2012- 2013 | Colombia | No explicit definition | Colombia has a personal data protection law and an International AI Council for Colombia. |
| 2018 | Mexico | No explicit definition | Mexico presented the first national AI strategy in the Latin American and Caribbean region in March 2018. The strategy focuses on ethics, governance, investment, and innovation. |
| 2019 | Brazil | The Brazilian AI Strategy defines AI as "a set of techniques that enable machines to perform tasks that, if performed by humans, would require intelligence". | The strategy focuses on AI research and development, ethics, and governance. |
| 2021 | Chile | No explicit definition | Chile has a national AI policy that focuses on ethics, governance, and innovation. |
| 2021 | Peru | No explicit definition | Peru has a national AI policy focusing on ethics, governance, and innovation. |

According to the same OECD Report, Annex A (2022) [30], some critical points regarding LAC nations' AI strategies include:

- Uruguay (AI Strategy for the Digital Government) focused on AI use for governmental administration, capacity building, responsible use, and digital citizenship.

- Peru (National Strategy for Artificial Intelligence) infrastructure, ethics, training, and economic models, among other strategies to incorporate AI in governmental operations. Renews strategy every two years after 2026.
- Mexico (IA-MX 2018) has an AI subcommittee that looks towards the best possible government and combined approaches. They have participated in technology-related working groups.
- Colombia (Digital Transformation and AI National Policy) also searches to include AI in government and governmental services.
- Chile (AI Action Plan and AI National Policy) looks for AI inclusion in sustainable development and other areas of well-being, as well as regulatory aspects.
- Brazil (Brazilian Artificial Intelligence Strategy) looks for AI use in government through regulation, use in public services, legislation, ethics, setting norms for use, and even international aspects.
- Argentina (AI National Plan) uses AI to address the needs of government and citizens and more efficient work at the State level.
- Uruguay (Agenda Uruguay Digital 2020) looks for responsible, equitable, and respectful AI strategies in the country, as well as the privacy of users [31, 32].
- Costa Rica lawmakers created an AI regulation bill draft using ChatGPT, a publicly available AI tool that became popular during the Winter of 2022. It was created to call for legislation regarding AI use and development [34].
- The Dominican Republic has a national posture about AI but has yet to create a route to enforce it.

The OECD report provides valuable guidance for LAC governments on maximising AI's positive impacts and minimising the negative ones. By following these recommendations, governments can help to ensure that AI is used for the benefit of all.

## 4. Democratizing AI Knowledge in LAC

The interest in AI has influenced several areas and, in many cases, directly impacting people's lives, such as healthcare. However, scaling invocations requires adaptation and acceptance, which is often challenging. Therefore, understanding which factors influence the adoption of new ideas plays an important role. Considering this, our network proposes to use Rogers' Diffusion of Innovation Theory [37], which aims to explain how new ideas or innovations can be adopted, such as (1) relative advantage, (2) compatibility, (3) complexity, (4) experimentability and (5) observability. We will also apply other measures for innovation adoption [38] (e.g., cost-efficacy, feasibility, perceived evidence, innovation fit with users' norms, relevance, and ease).

To democratise AI knowledge, we intend to maximise the dissemination of results and products among the target communities where our motivating issues arise, as well as the training of human resources. We know that to bring positive societal changes, we must educate young academics, health professionals, policymakers, journalists, and invested citizens. Therefore, our network proposes several approaches, such as (see Figure 1): (1) Engagement with underrepresented communities; (2) Special short courses; (3) Public awareness campaigns in AI; (4) Collaboration with Industry; (5) International collaboration; (6) Open-Source Initiatives; and (7) Community-based development.

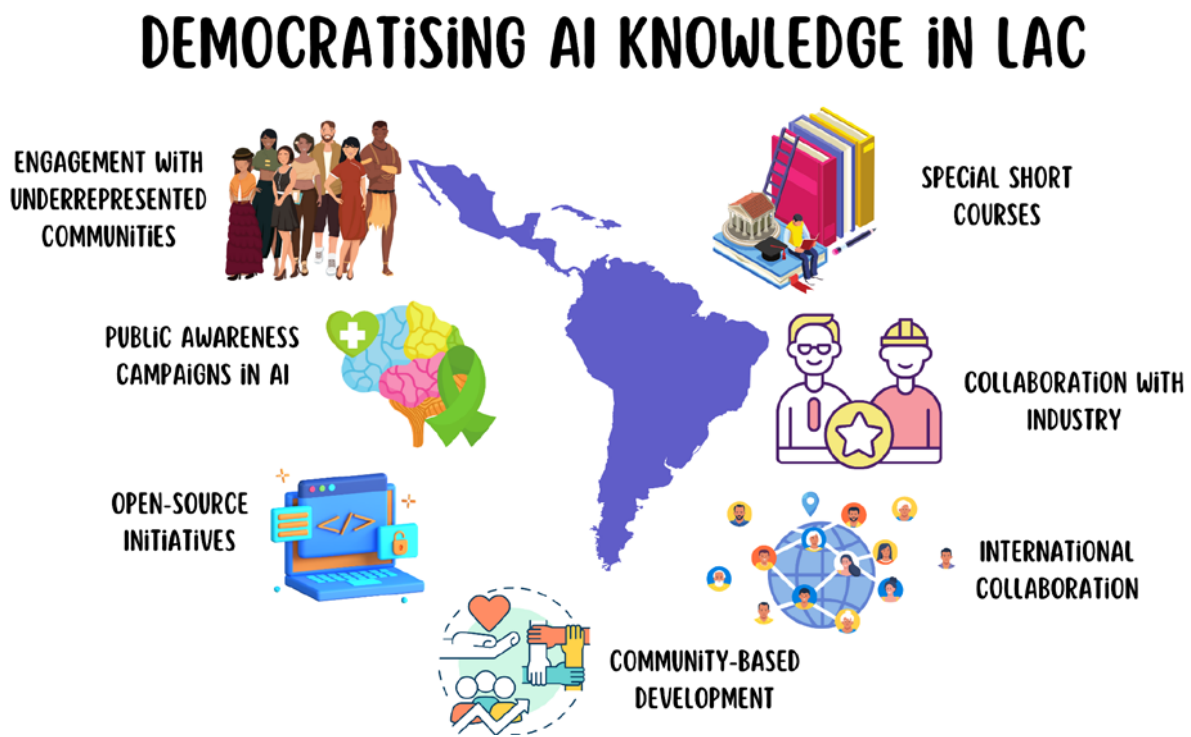

**Figure 1: Democratising AI Knowledge in Latino America and Caribbean countries**

Furthermore, our global south network intends to organise and support academic meetings to strengthen the collaboration of our multinational team and bridge it to local groups in Latin America. In these events, we will provide novel training sessions to environmental and healthcare professionals and students (e.g., tools developed for intervention design and combating misinformation). We will also encourage new partnerships around key challenges faced explicitly by AI initiatives in the Global South. We will make special efforts to finance the participation of promising students from other developing countries to provide them with novel training opportunities and possibilities for continued education within our post-graduate program. These initiatives will train professionals to work with AI involving pandemic preparedness and global governance in LAC and may serve as bridges between LAC communities to disseminate the new methods and tools.

## 5. Conclusion

As we face global challenges such as pandemics, AI emerges as a capable ally to mitigate the impact of these crises and diminish response time in real-time, promoting a more resilient world. However, the journey to harness AI's potential is fraught with ethical challenges. The democratisation of AI knowledge and tools in LAC, as proposed by our network, represents a fundamental step towards making AI accessible to a broader set of stakeholders, covering healthcare professionals, policymakers, and the general population. Therefore, this democratisation is a multifaceted effort that requires cooperation, innovation, and a firm commitment to ethical principles not only in Latin America and Caribbean countries but worldwide. By promoting international collaboration, disseminating knowledge, and adopting responsible AI practices, we can harness the transformative power of AI to build a more resilient and equitable world, prepared to face future challenges, whether pandemics or not. The LAC region plays a distinct role in this effort, contributing to the global discourse on AI governance and its conscious use in the service of humanity.

## Funding

This research is funded by Canada's International Development Research Centre (IDRC) (Grant No. 109981). UR is supported by the Deutsche Forschungsgemeinschaft under the NFDI4Microbiota consortium, grant number 460129525.

## Acknowledgment

We wish to acknowledge that this work is based on a report we originally submitted to the United Nations Call for Papers on Global AI Governance (https://www.un.org/techenvoy/ai-advisory-body). The insights and research contained herein were drawn from that report, and we are grateful for the opportunity to contribute to the ongoing discourse on global AI governance through this extended work.